\documentclass[
  journal=largetwo,
  manuscript=ORIGINAL-ARTICLE,
  year=2025,
  volume=37,
]{cup-journal}

\usepackage{amsmath}
\usepackage[nopatch]{microtype}
\usepackage{booktabs}
\usepackage{amssymb}
\usepackage{makecell}
\usepackage{stfloats}

\title{CFIS-YOLO: A Lightweight Multi-Scale Fusion Network for Edge-Deployable Wood Defect Detection}

\author{Jincheng Kang}
\affiliation{School of Information and Engineering, Minzu University of China, Beijing 100081, People’s Republic of China}

\author{Yi Cen}
\affiliation{School of Information and Engineering, Minzu University of China, Beijing 100081, People’s Republic of China}
\email[Yi Cen]{tsinge@muc.edu.cn}

\author{Yigang Cen}
\affiliation{School of Computer and Information Technology, Beijing Jiaotong University, Beijing 100044, People’s Republic of China}

\author{Ke Wang}
\affiliation{School of Information and Communication Engineering, Beijing University of Posts and Telecommunications, Beijing 100876, People’s Republic of China}

\author{Yuhan Liu}
\affiliation{Faculty of Science, The University of Hong Kong, Hong Kong SAR 999077, People’s Republic of China}

\keywords{Object detection, YOLOv10, Wood defects, Lightweight network, Edge deployment} 

\begin{document}

\begin{abstract}
Wood defect detection is critical for ensuring quality control in the wood processing industry. However, current industrial applications face two major challenges: traditional methods are costly, subjective, and labor-intensive, while mainstream deep learning models often struggle to balance detection accuracy and computational efficiency for edge deployment. To address these issues, this study proposes CFIS-YOLO, a lightweight object detection model optimized for edge devices. The model introduces an enhanced C2f structure, a dynamic feature recombination module, and a novel loss function that incorporates auxiliary bounding boxes and angular constraints. These innovations improve multi-scale feature fusion and small object localization while significantly reducing computational overhead. Evaluated on a public wood defect dataset, CFIS-YOLO achieves a mean Average Precision (mAP@0.5) of 77.5\%, outperforming the baseline YOLOv10s by 4 percentage points. On SOPHON BM1684X edge devices, CFIS-YOLO delivers 135 FPS, reduces power consumption to 17.3\% of the original implementation, and incurs only a 0.5 percentage point drop in mAP. These results demonstrate that CFIS-YOLO is a practical and effective solution for real-world wood defect detection in resource-constrained environments.
\end{abstract}

\section{Introduction}

As a crucial renewable resource, wood plays a crucial strategic role in industries such as construction, furniture manufacturing, and wood product processing \autocite{wlsd}. However, common surface defects on wood—such as knots, cracks, and wormholes—result in global wood utilization rates as low as 50\%-70\% \autocite{cwb4}. In high-end flooring production, even minor surface defects can lead to significant economic losses due to batch rejection \autocite{fdd31}. Traditional manual inspection methods suffer from low efficiency and strong subjectivity \autocite{cwb}, while non-destructive testing technologies based on laser scanning or X-rays, despite their ability to detect internal defects, face challenges in widespread adoption due to high equipment costs, operational complexity, and poor adaptability to complex textures \autocite{bpn}.

Breakthroughs in deep learning have enabled CNN-based object detection algorithms to provide efficient solutions for wood defect detection, gradually becoming a focus of research. Early studies focused on two-stage detection models (e.g., Faster R-CNN and its variants), achieving high-precision detection through region proposal mechanisms. For instance, \textcite{zou} achieved 67.8\% detection accuracy on an open-source wood dataset using an improved Faster R-CNN, though its complex architecture limited real-time performance. In contrast, single-stage models (e.g., YOLO series) have gained industrial preference due to their rapid inference capabilities. \textcite{2019} employed a 23-layer Tiny-YOLO network to detect wood pith, achieving 76.3\% accuracy on a custom dataset, although its scope was limited to pith detection.

Recent research has focused on multi-dimensional optimizations of YOLO architectures to address the diversity and scale variations of wood defects. At the feature extraction level, the Biformer attention mechanism has been widely adopted \autocite{bpn}\autocite{zheng2024}\autocite{cwb}. This module effectively captures both local and global semantic information by integrating Transformer and feature pyramid networks, enhancing perception of multi-scale defects. For feature fusion, \textcite{han2023} introduced BiFPN, optimizing multi-scale feature integration efficiency through weighted fusion mechanisms and improving small-defect detection accuracy. \textcite{yolov5s} adopted a Hybrid Spatial Pyramid Pooling Fusion (HSPPF) structure, significantly enhancing cross-level information integration by combining max-pooling and average-pooling operations while reducing information loss. Furthermore, loss function improvements have mitigated small-target localization biases and sample imbalance issues: \textcite{yolov5s} employed an improved Curved Efficient Complete Intersection over Union (CEIoU) loss function to enhance detection accuracy, while \textcite{wang2023} introduced Wise-IoU loss with dynamic weight optimization for better small-target detection.

Despite significant progress in existing research, wood defect detection still faces three core challenges:
\begin{enumerate}
    \item \textbf{Insufficient multi-scale feature fusion capability}: The diverse sizes and morphologies of wood defects make traditional upsampling methods (e.g., bilinear interpolation) inadequate. These methods rely on fixed kernel functions that struggle to adapt to the variability of wood textures and complexity of defect patterns.
    
    \item \textbf{Limited small object localization accuracy}: Tiny defects (e.g., wormholes or cracks) occupy minimal pixels in images. Traditional IoU loss becomes sensitive to bounding box offsets under the interference of annual ring textures.
    
    \item \textbf{Difficulty in balancing accuracy and efficiency during deployment}: Industrial scenarios demand strict real-time performance and low power consumption. However, most existing studies focus on theoretical performance without verifying deployment effectiveness on actual hardware.
\end{enumerate}

To resolve these challenges, a lightweight and efficient wood defect detection model, CFIS - YOLO, is proposed with three key innovations:

\begin{enumerate}
    \item \textbf{A lightweight Content-Aware ReAssembly of FEatures (CARAFE) operator}: By dynamically recombining kernels to enhance semantic alignment of multi-scale defect features, this module significantly reduces feature information loss compared to traditional methods. It significantly improves edge alignment accuracy for knots and cracks, effectively solving the multi-scale fusion problem.
    
    \item \textbf{FasterBlock module integrated into C2f structure}: Leveraging partial convolutions to reduce computational redundancy, this design maintains feature representation capability while lowering FLOPs. The optimized architecture achieves higher FPS inference speed, facilitating edge device deployment.
    
    \item \textbf{Inner-SIoU loss combining internal intersection-over-union and angle awareness}: This novel loss function integrates auxiliary bounding boxes and angular constraints to optimize small object localization. It reduces positioning error for micro-defects, addressing the critical challenge of limited small target detection accuracy.
\end{enumerate}

The results of the experiment reveal that our model attains 77.5\% mAP@0.5 on public wood defect datasets, marking a 4 percentage point improvement over the baseline YOLOv10s. Deployment tests on SOPHON BM1684X edge devices show power consumption reduced to 17.3\% of original levels, with only 0.5 percentage point mAP compromise. This study proposes an efficient and robust methodology for intelligent upgrading in the wood processing industry by integrating innovative feature fusion, loss function design, and network optimization.

\section{Methodology}
\subsection{Overall architecture}

YOLOv10\autocite{yolov10} is the new generation of YOLO series, inheriting the advantages of previous YOLO models while introducing several improvements that further enhance performance. One of its most significant innovations is the introduction of the Non-Maximum Suppression (NMS)-free consistent dual assignment strategy. This strategy eliminates the dependency on NMS post-processing by combining dual label assignment with consistent matching metrics, thereby achieving more efficient inference\autocite{ali2024yolo}.

\begin{figure}[hbt!]
\centering
\includegraphics[width=0.99\linewidth]{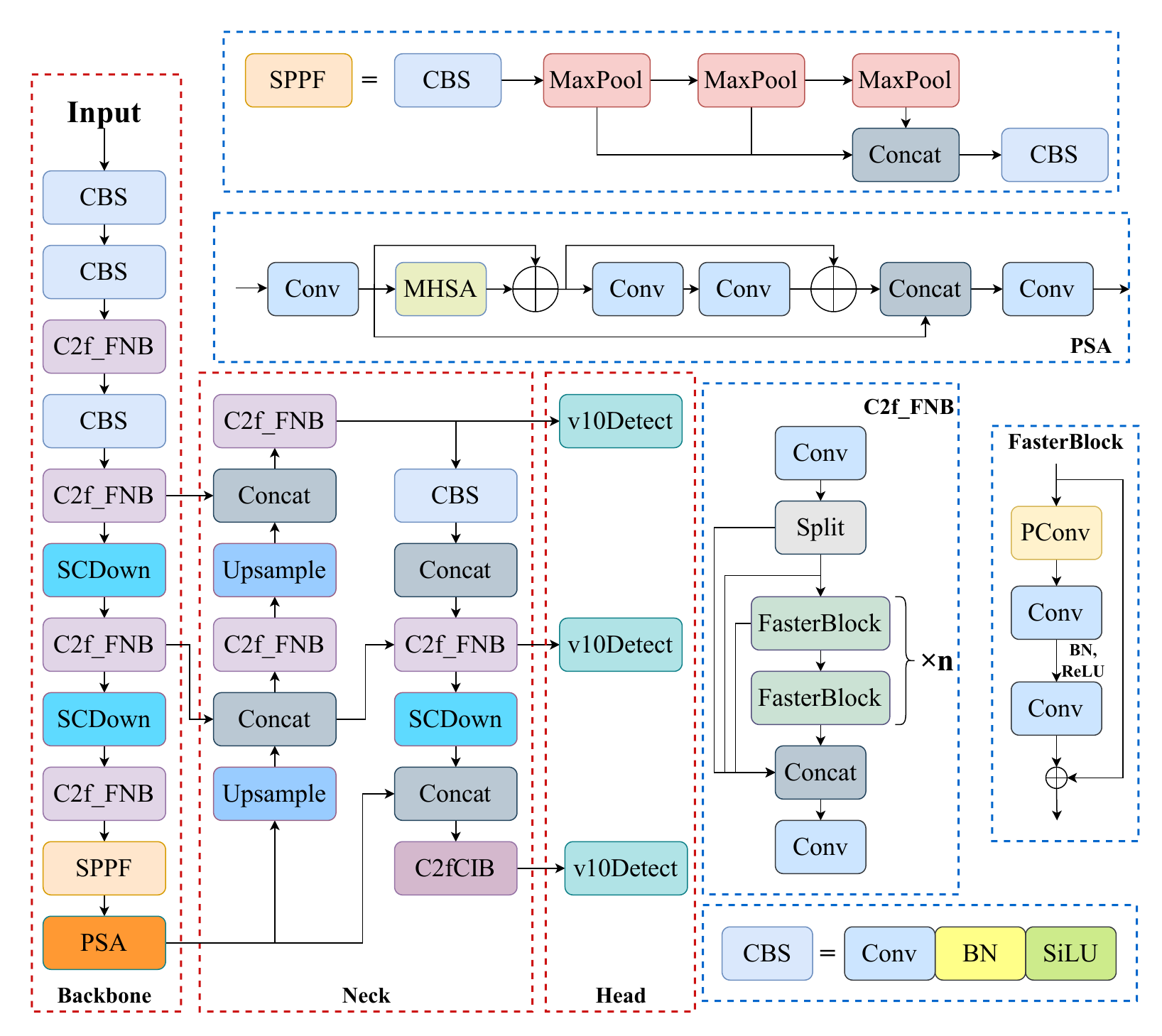}
\caption{Structure of the proposed CFIS-YOLO.}
\label{yolov10-wood}
\end{figure}

Based on the YOLOv10 benchmark architecture, this study proposes the CFIS-YOLO lightweight detection framework with innovative improvements. As shown in Figure \ref{yolov10-wood}, the improved network retains the backbone network, feature fusion module, and detection head structure, but significantly enhances wood defect detection performance through three core improvements: 1) Replacing the upsampling method with the lightweight dynamic upsampling operator CARAFE to enhance multi-scale defect representation through adaptive feature reorganization; 2) Integrating FasterBlock structure into the C2f module, adopting partial convolution (PConv) to reduce redundant computation; 3) Designing the Inner-SIoU loss function that combines internal intersection-over-union and angular constraints to optimize small object localization accuracy. The following sections will elaborate on the design principles and implementation details of each improved module.

\subsection{Lightweight Upsampling Operator CARAFE}

Feature upsampling is a critical operation in object detection tasks. YOLOv10 employs nearest neighbor interpolation as its upsampling method\autocite{zou2025}, which assigns the gray value of the nearest neighboring pixel in the transformed image to the original pixel. While computationally simple, this method determines the upsampling kernel solely based on pixel spatial positions without utilizing semantic information from feature tensors, and typically has limited receptive fields. The CARAFE upsampling operator\autocite{carafe} overcomes these limitations by achieving larger receptive fields during feature rearrangement, dynamically performing upsampling operations based on input data, and maintaining low computational cost.

\begin{figure}[hbt!]
\centering
\includegraphics[width=0.99\linewidth]{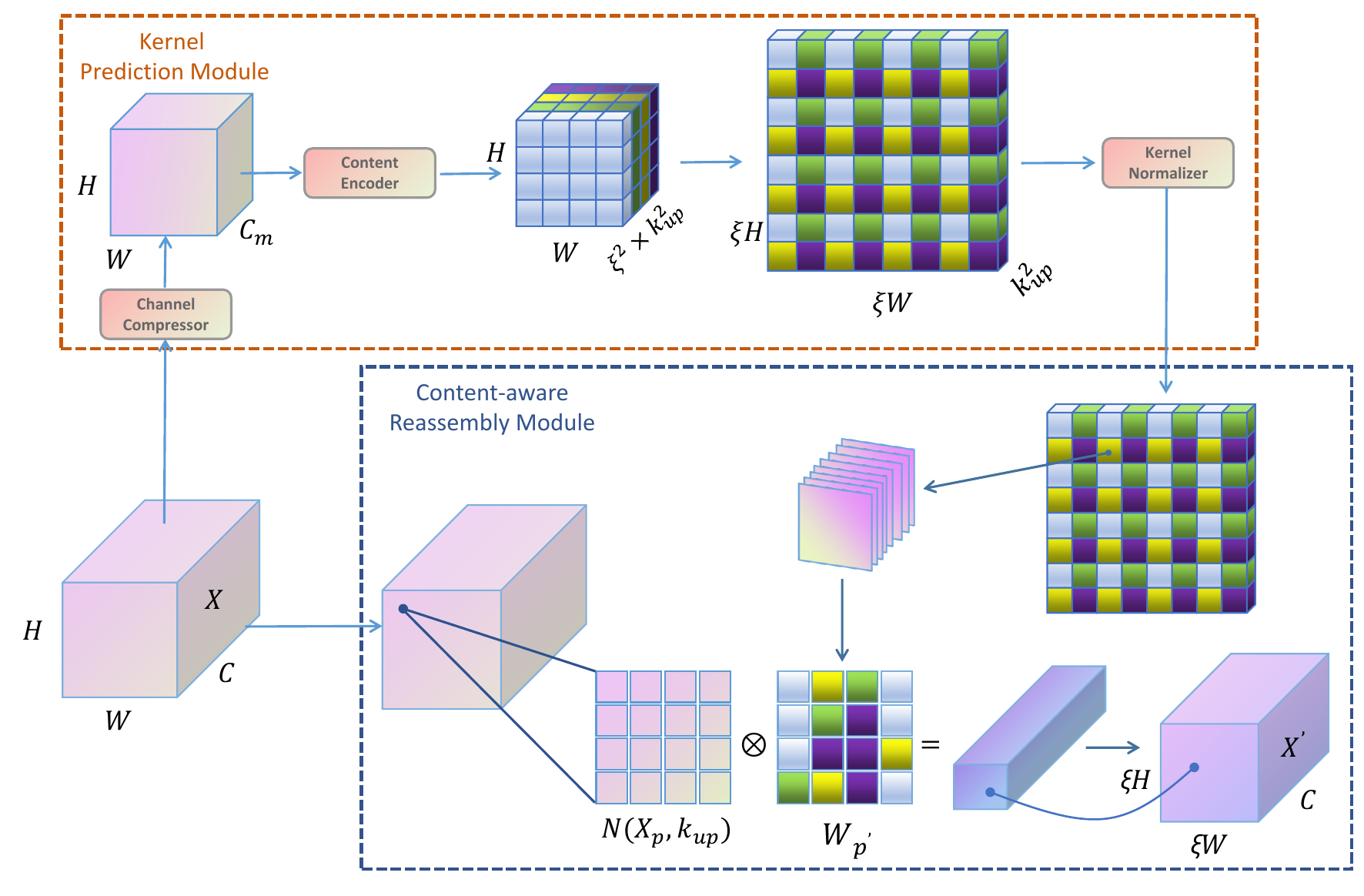}
\caption{Upsampling process of the CARAFE operator.}
\label{carafe}
\end{figure}

As shown in Figure \ref{carafe}, CARAFE consists of two components: the kernel prediction module and the content-aware reassembly module.

Given input tensor $X \in \mathbb{R}^{C \times H \times W}$, CARAFE upsamples to $X' \in \mathbb{R}^{C \times \xi H \times \xi W}$ via integer scaling factor $\xi$. Each output coordinate $p'=(u',v')$ in $X'$ maps to input coordinate $p=(u, v)$ in $X$, where $u = \lfloor u'/\xi \rfloor$ and $v = \lfloor v'/\xi \rfloor$.

\subsubsection{The Kernel Prediction Module}

The proposed module generates the reassembly kernel $W_{p'}$ for each target spatial location $p'$ by leveraging the local neighborhood $N(X_p, k_{\text{encoder}})$ centered at position $p$ within the input feature tensor $X$. This mapping procedure can be formally expressed as:
\begin{equation}
W_{p'} = \sigma(N(X_p, k_{\text{encoder}})) 
\end{equation}
where $\sigma$ denotes the kernel prediction module, \( X_p \) denotes the coordinate of a pixel on \( X \), and \( k_{\text{encoder}} = k_{\text{up}} - 2 \) represents the size of the convolution kernel.

The kernel prediction module consists of three operations: channel compression, content encoding, and kernel normalization. These steps are explained below:
\begin{enumerate}
\item 
\textbf{Channel compression}: A 1×1 convolution layer reduces input channels from $C$ to $C_m$, decreasing parameters and computational overhead in downstream processes.

\item 
\textbf{Content encoding}: A $k_{\text{encoder}}$-sized convolutional layer generates the reassembly kernel with parameter count $k_{\text{encoder}}^2 \times C_m \times C_{\text{up}}$. Increasing $k_{\text{encoder}}$ expands the receptive field to capture broader contextual information, but computational cost scales quadratically with $k_{\text{encoder}}$, requiring a trade-off between performance and efficiency.

\item 
\textbf{Kernel normalization}: Each $k_{\text{up}} \times k_{\text{up}}$ reassembly kernel is normalized by softmax prior to application, enforcing unit-sum constraints for soft local selection. This normalization preserves feature tensor means through the CARAFE process.
\end{enumerate}

\subsubsection{The Content-aware Reassembly Module}

The module employs predicted kernel $W_{p'}$ to reconstruct features from neighborhood $N(X_p, k_{\text{up}})$ in input tensor $X$, generating output values at position $p'$ in $X'$. This operation is formally defined as:
\begin{equation}
X'_{p'} = \phi(N(X_p, k_{\text{up}}), W_{p'})
\end{equation}
where $\phi$ represents the content-aware reassembly module, and $k_{\text{up}}$ denotes the kernel size for reassembly.

For each target position $p'$, the content-aware reassembly module utilizes kernel $W_{p'}$ and function $\phi$ to redistribute features from the local region $N(X_p, k_{\text{up}})$ centered at $p$. With $r = \lfloor k_{\text{up}}/2 \rfloor$, this reassembly is defined as:

\begin{equation}
X'_{p'} = \sum_{n=-r}^{r} \sum_{m=-r}^{r} W_{p'}(n,m) \cdot X(u+n,v+m)
\end{equation}
where $X'_{p'}$ denotes the output feature at position $p'$. The kernel $W_{p'}$ ensures pixels within the local region $N(X_p, k_{\text{up}})$ contribute to $X'_{p'}$ based on feature content, rather than mere spatial distance. This content-aware weighting can enhance the semantic richness of the reassembled features $X'$ compared to the input $X$, as relevant information receives greater emphasis.

\subsection{FasterNet Network and C2f\_FNB Module}

\begin{figure}[hbt!]
\centering
\includegraphics[width=0.75\linewidth]{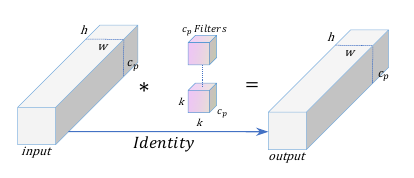}
\caption{Partial Convolution.}
\label{pconv}
\end{figure}

FasterNet\autocite{faster} is a lightweight neural network featuring Partial Convolution (PConv). PConv leverages feature map redundancy by applying standard convolution to only a subset of input channels, leaving the remaining channels unchanged. Figure \ref{pconv} illustrates the working principle of PConv.  

For contiguous memory access patterns, the first/last \( c_p \) contiguous channels serve as representatives for the entire feature tensor. Assuming equal input/output channels (\(c\)), the computational complexity of PConv is:
\begin{equation}
h \times w \times k^2 \times {c_p}^2
\end{equation}
compared to standard convolution (\( h \times w \times k^2 \times c^2 \)), setting channel ratio \( r = c_p/c = 1/4 \) yields a 16× FLOPs reduction. 

Inspired by the FasterNet network, this study replaces the C2f module’s bottleneck with a FasterBlock, yielding the enhanced C2f\_FNB module. Specifically, the input feature tensor is first processed by a 1×1 convolution, doubling the number of output feature tensor channels, which enhances the model's feature representation capability. Inside the C2f\_FNB module, the input feature tensor is split into two parts: one enters the FasterBlock, and the other is directly used for subsequent concatenation. The split feature tensors are processed layer by layer through multiple FasterBlocks to extract deeper features. The incorporated PConv reduces redundant computation and memory access while maintaining effective spatial feature extraction. Subsequently, the outputs of all FasterBlocks and the previously split feature tensors are concatenated to increase feature diversity. Finally, a convolutional layer compresses the concatenated feature tensor channels to the required output channels to meet the demands of subsequent processing.  

This improvement not only maintains high FLOPS but also significantly reduces FLOPs, effectively minimizing unnecessary computation and memory usage\autocite{PGY}. As a result, it reduces model size and accelerates inference time, making it more favorable for edge deployment. The structure of the C2f\_FNB module is shown on the right side of Figure \ref{yolov10-wood}.

\subsection{Inner-SIoU Bounding Box Loss Function}

The default bounding box loss in YOLOv10 is CIoU. However, its formulation, despite incorporating overlap, center distance, and aspect ratio metrics, omits the angular difference, which can lead to slower convergence and reduced prediction accuracy. To address this, this study replaces CIoU with the Inner-SIoU\autocite{inner} bounding box loss function, aiming to compensate for CIoU's limitations by incorporating angular loss between predicted and ground-truth boxes.

\begin{enumerate}

\item \textbf{Auxiliary Bounding Box Construction}\\
Let $\mathbf{t} = (x_t, y_t, w_t, h_t)$ and $\mathbf{a} = (x_a, y_a, w_a, h_a)$ represent ground-truth and anchor boxes, respectively. The scaling factor $\gamma \in [0.5, 1.5]$ controls auxiliary box dimensions.  Auxiliary coordinates (\text{tl} = top-left, \text{br} = bottom-right) are derived as:
\begin{align}
\hat{\mathbf{t}}_{\text{tl}} &= \left(x_t - \frac{\gamma w_t}{2}, y_t - \frac{\gamma h_t}{2}\right) \\
\hat{\mathbf{t}}_{\text{br}} &= \left(x_t + \frac{\gamma w_t}{2}, y_t + \frac{\gamma h_t}{2}\right) \\
\hat{\mathbf{a}}_{\text{tl}} &= \left(x_a - \frac{\gamma w_a}{2}, y_a - \frac{\gamma h_a}{2}\right) \\
\hat{\mathbf{a}}_{\text{br}} &= \left(x_a + \frac{\gamma w_a}{2}, y_a + \frac{\gamma h_a}{2}\right)
\end{align}

\item \textbf{Intersection Area Computation:}
\begin{equation}
\begin{split}
\mathcal{A}_{\cap} = \bigg[ & \min(\hat{\mathbf{t}}_{\text{br}}^x, \hat{\mathbf{a}}_{\text{br}}^x) - \max(\hat{\mathbf{t}}_{\text{tl}}^x, \hat{\mathbf{a}}_{\text{tl}}^x) \bigg]^+ \\
& \cdot \bigg[ \min(\hat{\mathbf{t}}_{\text{br}}^y, \hat{\mathbf{a}}_{\text{br}}^y) - \max(\hat{\mathbf{t}}_{\text{tl}}^y, \hat{\mathbf{a}}_{\text{tl}}^y) \bigg]^+
\end{split}
\end{equation}

\item \textbf{Union Area Formulation:}
\begin{equation}
\mathcal{A}_{\cup} = \gamma^2(w_t h_t + w_a h_a) - \mathcal{A}_{\cap}
\end{equation}

\item \textbf{Inner-IoU Metric:}
\begin{equation}
\mathrm{IoU}_{\text{inner}} = \frac{\mathcal{A}_{\cap}}{\mathcal{A}_{\cup}}
\end{equation}
\end{enumerate}

Subsequently, SIoU \autocite{siou} is incorporated. This method employs an angle alignment mechanism that optimizes bounding box regression by redefining angular penalty metrics, enabling predicted boxes to rapidly align with the nearest coordinate axis under vector angle guidance. By introducing distance loss and shape loss, the bounding box regression is optimized across three dimensions: directional alignment, positional offset, and size discrepancy.

\begin{enumerate}
    \item \textbf{Distance loss:} Based on angular alignment, the penalty weight for center distance is dynamically adjusted. When angular deviation is large, the distance penalty weight decreases, prioritizing directional correction before positional optimization. Denoted as \(\Delta_{dist}\).

    \item \textbf{Shape loss:} Penalizes aspect ratio differences between predicted and ground-truth boxes, amplifying dimensional errors through exponential functions to enhance sensitivity to significant size discrepancies. Denoted as \(\Delta_{shape}\).
\end{enumerate}

The SIoU loss function is defined as, where $\lambda_1$ and $\lambda_2$ are parameters:
\begin{equation}
L_{\text{SIoU}} = 1 - \text{IoU} + \lambda_{1}\cdot\Delta_{dist} + \lambda_{2}\cdot\Delta_{shape}
\label{eq:siou}
\end{equation}


The final Inner-SIoU loss function is defined as:
\begin{equation}
L_{\text{Inner-SIoU}} = L_{\text{SIoU}} + (\text{IoU} - \text{IoU}_{\text{inner}})
\end{equation}

\section{Experimental Setup and Result Analysis}
\subsection{Experimental Environment and Datasets}
\subsubsection{Experimental Environment}

\textbf{Experimental Setup.} The hardware configuration comprises an Intel Xeon Platinum 8369B processor paired with an NVIDIA A10 GPU (24GB VRAM), 32GB system memory, and Ubuntu 20.04 LTS. Our implementation leverages the PyTorch 2.0.1 framework with CUDA 11.7 acceleration and Python 3.8.16 interpreter. 

\textbf{Training Protocol.} All networks undergo 300 training epochs with $640 \times 640$ resolution inputs. We adopt stochastic gradient descent (SGD) optimization with momentum $\mu = 0.937$ and batch size 32. The learning rate follows a decay schedule from $\eta_{\text{initial}} = 0.01$ to $\eta_{\text{final}} = 0.001$. For sample selection, we implement IoU-based thresholding with $\tau = 0.5$ to differentiate positive/negative instances.Additionally, the parameters in equation (\ref{eq:siou}) are set as: $\lambda_1 = \lambda_2 = 0.5$.

\subsubsection{Datasets}

\begin{figure}[hbt!]
\centering
\includegraphics[width=0.99\linewidth]{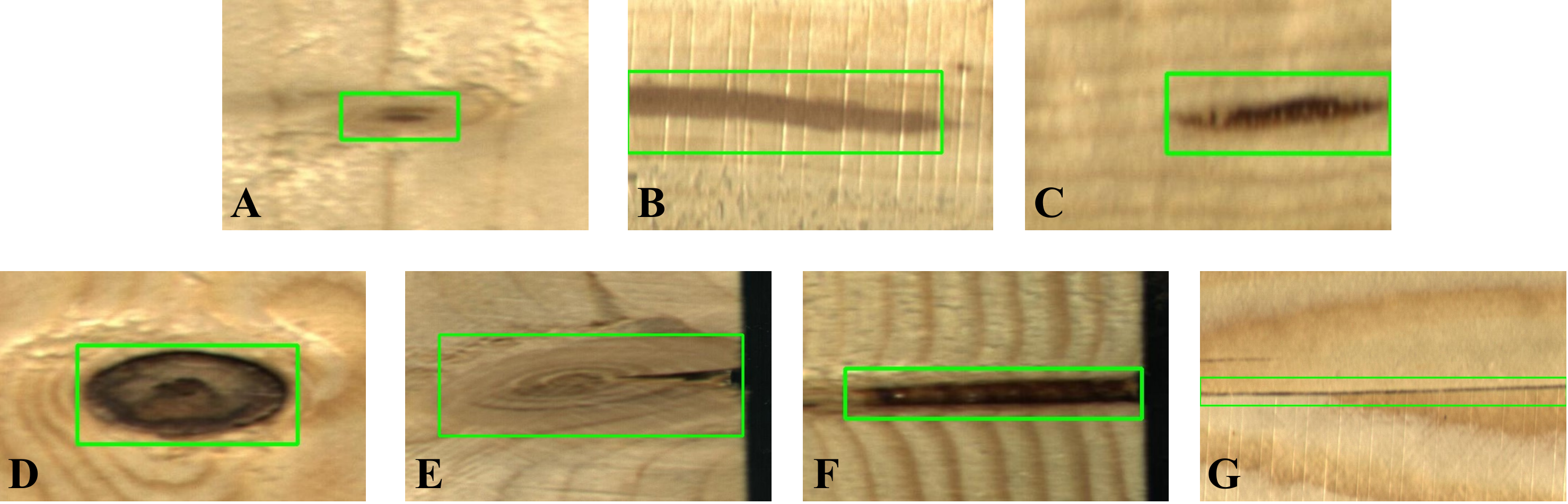}
\caption{The image displays typical wood defect samples within the dataset, including: (A) Live Knot, (B) Marrow, (C) Resin, (D) Dead Knot, (E) Knot with Crack, (F) Knot Missing, (G) Crack.}
\label{sjj}
\end{figure}

\begin{figure}[hbt!]
\centering
\includegraphics[width=0.8\linewidth]{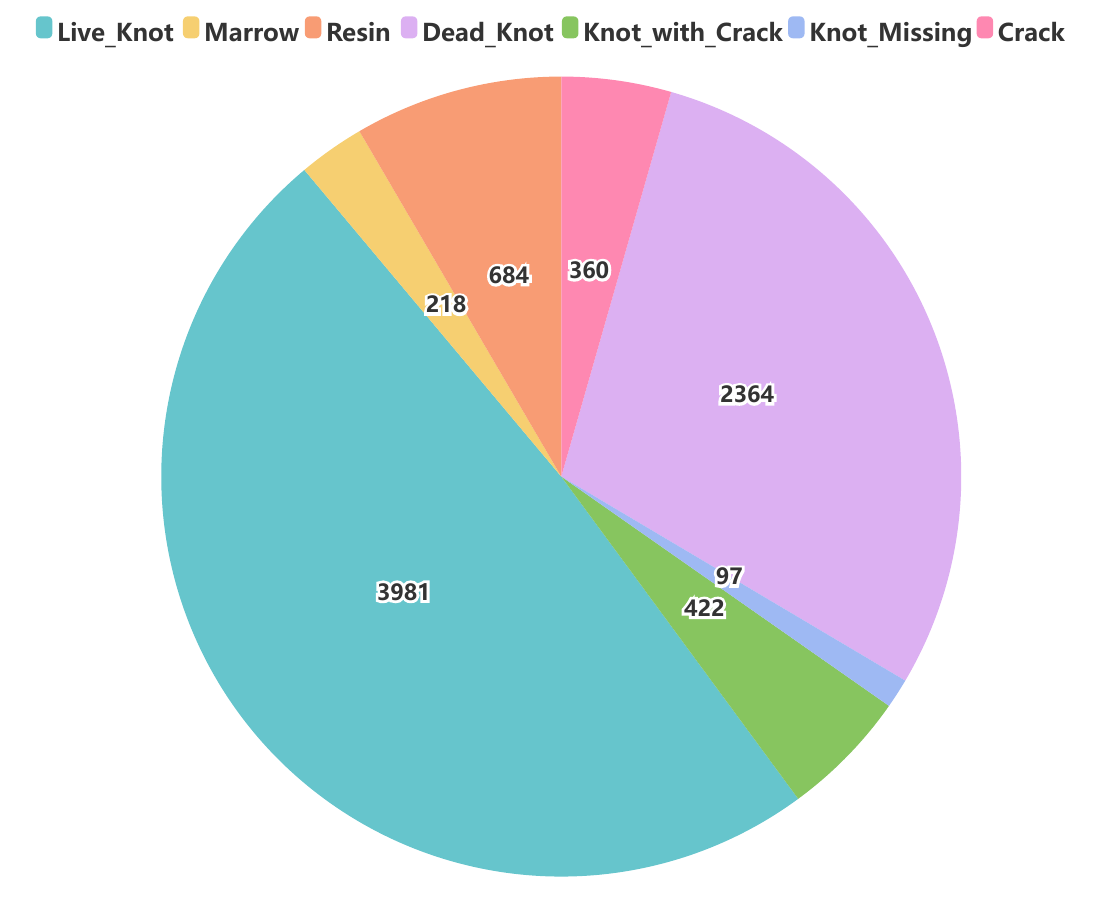}
\caption{Number of various annotations in the dataset.}
\label{labels}
\end{figure}

This study utilizes the wood surface defect dataset released by VŠB-TUO\autocite{sjj}. The original dataset contains 20,275 images with a resolution of 2800 × 1024, including 1,992 defect-free images and 18,283 images with one or multiple surface defects. The dataset covers ten types of wood defects: Live Knot, Dead Knot, Quartzity,  Knot with crack, Knot missing, Crack, Overgrown, Resin, Marrow, and Blue stain. Excluding three rarely occurring defects (Quartzity, Blue stain, and Overgrown), we used 3,465 images partitioned into training, test, and validation sets at an 8:1:1 ratio. Figure \ref{sjj} shows example images of these wood defects, while Figure \ref{labels} visually presents the annotation details in the dataset.

\subsection{Performance Metrics}

Model performance evaluation uses Precision, Recall, Average Precision (AP), mAP, Params, and Frames Per Second (FPS) as evaluation metrics. The calculation formulas for P, R, AP, and mAP are as follows:

\begin{equation}
P = \frac{Y_1}{Y_1 + N_1}
\end{equation}

\begin{equation}
R = \frac{Y_1}{Y_1 + N_2}
\end{equation}

\begin{equation}
\text{AP} = \int_{0}^{1} P(R) \, dR
\end{equation}

\begin{equation}
\text{mAP} = \frac{1}{n}{\sum_{i=0}^{n} \text{AP}_{(i)}}
\end{equation}


In Equations (14)-(17), \( Y_1 \), \( N_1 \), and \( N_2 \) represent true positives, false positives, and false negatives, respectively. \( n \) denotes the total number of classes, \( \text{AP}_{(i)} \) represents the AP value for the \( i \)-th class, and \( P(R) \) represents the probability of correct object detection corresponding to recall \( R \).

\subsection{Ablation Experiment}

To systematically evaluate the independent contributions and synergistic effects of the proposed CARAFE, C2f\_FNB, and Inner-SIoU modules, this study incrementally integrates each module into a baseline model and compares different combinations in terms of precision (P), recall (R), mean average precision (mAP@0.5), and parameter count on the validation set. The experimental results are shown in Table \ref{xrsy}, and the specific analysis is as follows.

\begin{table}[htbp]
\small
\centering
\setlength{\tabcolsep}{3pt}
\caption{Comparison of experimental results.}
\label{xrsy}
\begin{tabular}{cccccccc}
\toprule
ID & 
\makecell[c]{CARAFE} & 
\makecell[c]{C2f\_FNB} & 
\makecell[c]{Inner-\\SIoU} & 
\makecell[c]{P\\(\%)} & 
\makecell[c]{R\\(\%)} & 
\makecell[c]{mAP@0.5\\(\%)} & 
\makecell[c]{Params\\($10^6$)} \\
\midrule
1 & $\times$ & $\times$ & $\times$ & 73.8 & 70.4 & 73.5 & 8.07 \\
2 & $\checkmark$ & $\times$ & $\times$ & 79.1 & 70.3 & 75.3 & 8.23 \\
3 & $\times$ & $\checkmark$ & $\times$ & 72.3 & 73.9 & 75.6 & 7.01 \\
4 & $\times$ & $\times$ & $\checkmark$ & 74.5 & 73.2 & 75.0 & 8.07 \\
5 & $\checkmark$ & $\checkmark$ & $\times$ & 74.5 & 71.1 & 75.1 & 7.17 \\
6 & $\checkmark$ & $\times$ & $\checkmark$ & 71.1 & 72.6 & 74.5 & 8.23 \\
7 & $\times$ & $\checkmark$ & $\checkmark$ & 78.3 & 70.7 & 76.1 & 7.01 \\
8 & $\checkmark$ & $\checkmark$ & $\checkmark$ & 76.3 & 73.6 & 77.5 & 7.17 \\
\bottomrule
\end{tabular}
\end{table}

\subsubsection{Independent Effects of Modules}
\begin{enumerate}
    \item \textbf{CARAFE Module}: When CARAFE is introduced alone (Experiment 2), mAP@0.5 increases from 73.5\% of the baseline model to 75.3\% (+1.8 percentage points), with only a 0.16 million increase in parameters (from 8.07M to 8.23M). This indicates that CARAFE's dynamic convolution kernel upsampling mechanism significantly enhances the model's ability to capture detailed features with controllable computational overhead.
    
    \item \textbf{C2f\_FNB Module}: When C2f\_FNB is used alone (Experiment 3), mAP@0.5 improves to 75.6\% (+2.1 percentage points), while the parameter count decreases by 1.06 million (from 8.07M to 7.01M). Its lightweight design and efficient computational architecture effectively maintain detection accuracy while reducing model complexity.
    
    \item \textbf{Inner-SIoU Module}: When only Inner-SIoU is integrated (Experiment 4), mAP@0.5 reaches 75.0\% (+1.5 percentage points), with the parameter count remaining the same as the baseline model (8.07M). This module optimizes the bounding box regression process through geometric constraints, demonstrating its potential for improving localization accuracy.
\end{enumerate}

\subsubsection{Synergistic Effects of Module Combinations}
\begin{enumerate}
    \item \textbf{CARAFE + C2f\_FNB}: When CARAFE and C2f\_FNB are used together (Experiment 5), mAP@0.5 is 75.1\% (+1.6 percentage points), with the parameter count reduced to 7.17M (-0.9M). However, its performance is slightly lower than using C2f\_FNB alone (75.6\%), indicating partial functional overlap in feature representation optimization between the two modules.
    
    \item \textbf{CARAFE + Inner-SIoU}: This combination (Experiment 6) results in an mAP@0.5 of 74.5\% (+1.0 percentage points), but precision (P) decreases by 2.7 percentage points. This suggests that noise introduced by feature upsampling may conflict with the bounding box regression objectives, causing fluctuations in precision.
    
    \item \textbf{C2f\_FNB + Inner-SIoU}: This combination (Experiment 7) achieves the best performance among dual-module integrations, with mAP@0.5 reaching 76.1\% (+2.6 percentage points) and a parameter count of 7.01M. The efficient computation of C2f\_FNB and the regression precision optimization of Inner-SIoU demonstrate strong synergy, validating their complementary roles in model lightweighting and detection accuracy.
\end{enumerate}

\subsubsection{Performance of Full Module Integration}
When CARAFE, C2f\_FNB, and Inner-SIoU are integrated together (Experiment 8), the model achieves optimal performance: mAP@0.5 reaches 77.5\% (+4.0 percentage points), with the parameter count reduced to 7.17M (-0.9M). Despite partial redundancy between CARAFE and C2f\_FNB, the synergy of the three modules covers three critical dimensions: feature enhancement, computational efficiency, and localization optimization, significantly improving the overall performance of the detection framework.

\begin{figure}[hbt!]
\centering
\includegraphics[width=0.99\linewidth]{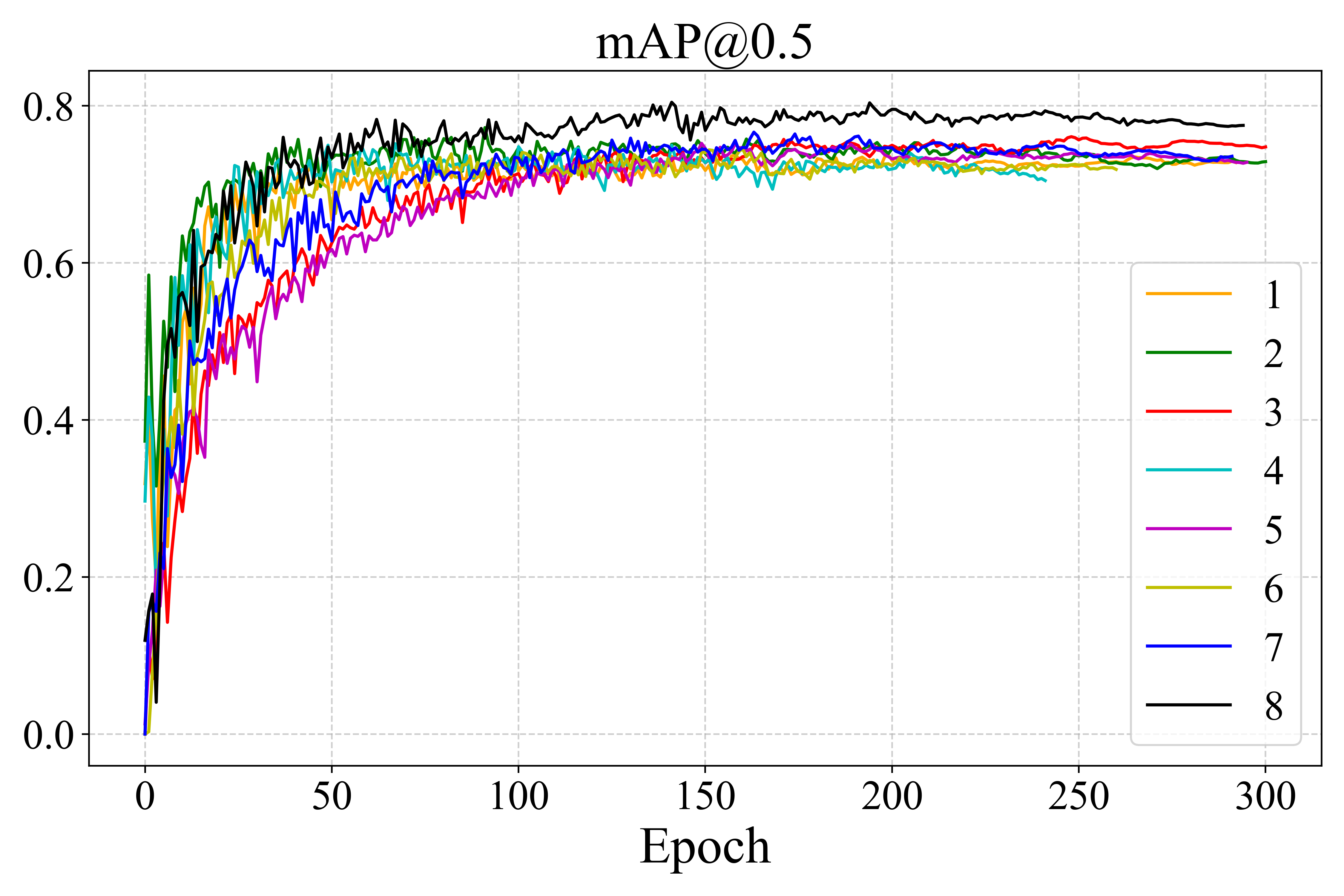}
\caption{The mAP curve graphs of the models.}
\label{xrsytu}
\end{figure}

\subsubsection{Discussion and Conclusion}
The ablation experiments demonstrate that the full integration of CARAFE, C2f\_FNB, and Inner-SIoU achieves optimal performance on the validation set, reducing the parameter count while improving mAP by 4 percentage points compared to the baseline model. Figure \ref{xrsytu} shows the mAP curves of the models during training. As illustrated, the synergy of the three modules significantly mitigates the limitations of individual modules: CARAFE enhances detailed feature representation through dynamic upsampling, C2f\_FNB reduces computational redundancy and accelerates inference, and Inner-SIoU optimizes the geometric constraints of bounding box regression. This multi-level optimization strategy not only balances accuracy and efficiency but also enhances model stability across different training phases, validating the robustness of the proposed improvements in wood defect detection tasks.

\subsection{Comparative Experiment}

To validate the performance advantages of our model compared to other mainstream object detection models, this section conducts comparative experiments using multiple methods. To ensure fairness and rigor, all comparisons are performed under consistent experimental environments and evaluation metrics.  
Table \ref{dbsy} presents the performance comparison between our model and current mainstream object detection models, with evaluation metrics including precision (P\%), recall (R\%), mean Average Precision at IoU threshold 0.5 (mAP@0.5 (\%)), and model parameter count (Params (\(10^6\))). Among them, the data of Faster R-CNN and RT-DETR are cited from the research of \textcite{drr}. They also used the VŠB-TUO dataset, but randomly selected 3,600 images from it.

Experimental results demonstrate that our model outperforms comparative models across all key metrics. Specifically, our model achieves a precision of 76.3\%, which is 2.5 percentage points higher than YOLOv10s (73.8\%); a recall of 73.6\%, surpassing YOLOv8s (71.7\%) by 1.9 percentage points; and an mAP@0.5 of 77.5\%, exceeding YOLOv8s (74.1\%) by 3.4 percentage points. This conclusively proves the significant advantages of our model in detection accuracy and comprehensiveness. Additionally, our model has 7.17M parameters, slightly higher than YOLOv5s (7.13M) and YOLOv7-tiny (6.08M) but significantly lower than YOLOv8s (11.2M) and YOLOv10s (8.07M), demonstrating effective control of computational complexity while maintaining high performance.

\begin{table}[htbp]
\small
\centering
\setlength{\tabcolsep}{3pt}
\caption{Performance comparison of different object detection models.}
\label{dbsy}
\begin{tabular}{cccccc}
\toprule
Method & P (\%) & R (\%) & mAP@0.5 (\%) & Params ($10^6$) \\
\midrule
Faster R-CNN & 43.8 & 70.2 & 59.2 & / \\
RT-DETR & 69.0 & 61.2 & 64.8 & / \\
YOLOv5s & 67.9 & 67.3 & 69.0 & 7.13 \\
YOLOv7-tiny & 73.7 & 66.1 & 71.0 & 6.08 \\
YOLOv8s & 72.9 & 71.7 & 74.1 & 11.2 \\
YOLOv9s & 73.4 & 69.1 & 71.6 & 7.1 \\
YOLOv10s & 73.8 & 70.4 & 73.5 & 8.07 \\
Ours & 76.3 & 73.6 & 77.5 & 7.17 \\
\bottomrule
\end{tabular}
\end{table}

\begin{figure*}[ht]
\centering
\includegraphics[width=0.87\textwidth]{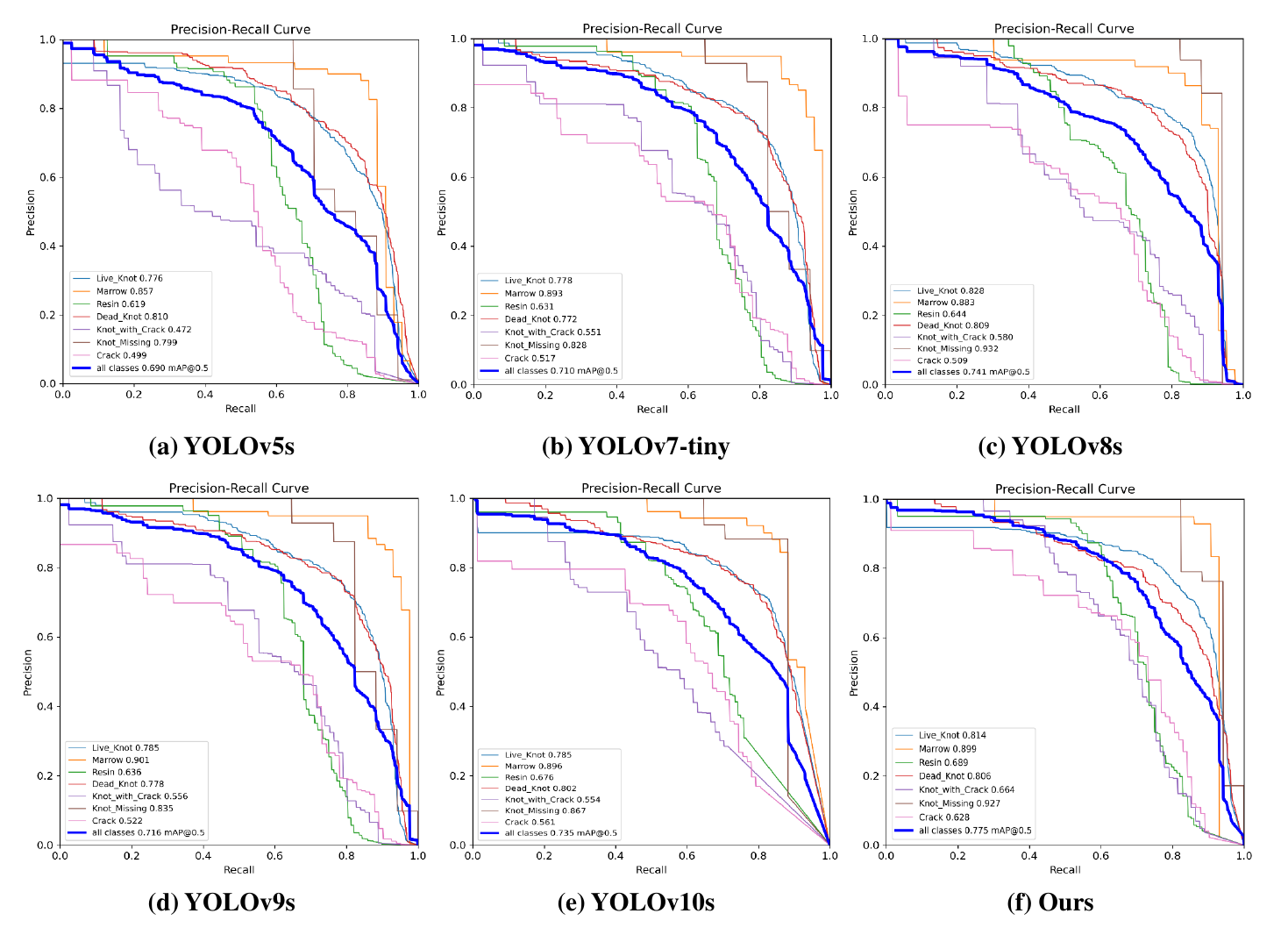}
\caption{PR curves of different YOLO models, AUC is higher for better detection performance.}
\label{pr}
\end{figure*}

Figure \ref{pr} illustrates the precision-recall (PR) curves of our model and comparative models. The PR curve plots the relationship between precision and recall at different thresholds, reflecting the performance trade-off in object detection tasks. The area under the PR curve (AUC-PR), a core metric for evaluating model capability, indicates stronger performance when higher. Our model exhibits significantly higher precision in high-recall regions compared to others, demonstrating superior robustness and detection capability in complex scenarios. Moreover, the less fluctuating nature of the PR curve for our model signifies its robustness, thereby reinforcing its credibility for real-world usage.

\begin{figure}[h]
\centering
\includegraphics[width=0.99\linewidth]{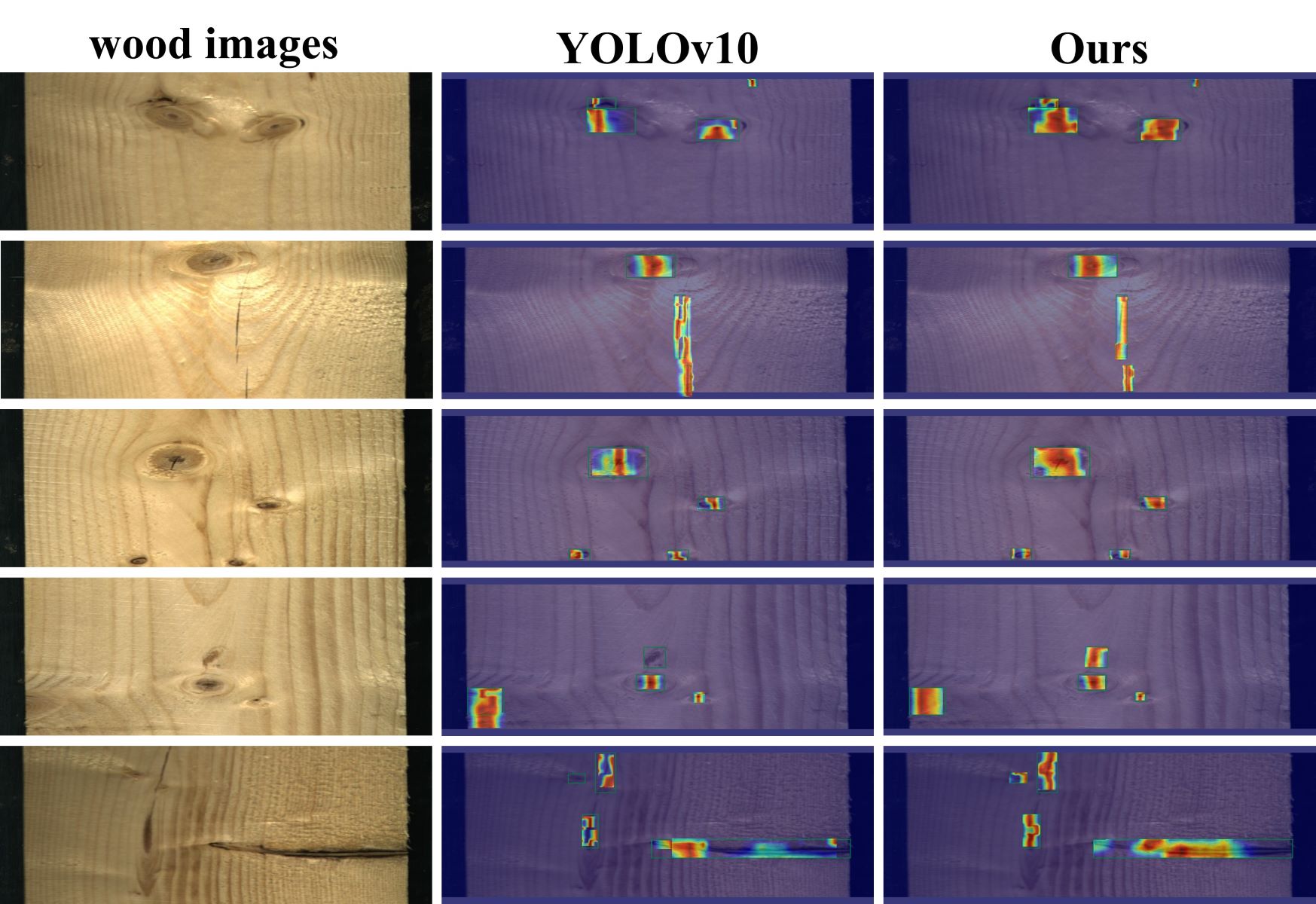}
\caption{Feature visualization of YOLOv10 and CFIS-YOLO.}
\label{rlt}
\end{figure}

Finally, Figure \ref{rlt} shows the heatmap visualizations of our model and YOLOv10, intuitively presenting their detection and localization capabilities across different target regions. The heatmaps, generated using Grad-CAM, highlight image regions where the model focuses, with color intensity indicating attention strength. Our model demonstrates more effective feature capture in critical target regions, such as small and multi-scale objects. For instance, in the second image containing two adjacent slender defects, YOLOv10 misidentifies them as a single defect, whereas our model correctly recognizes both. In the fourth image, while YOLOv10 detects the defect in the lower-left corner, it focuses on a broader area, whereas our model precisely localizes the defect region. These observations confirm that our model focuses more accurately on target areas, significantly reducing attention to background noise and further validating its superiority in complex scenarios.

\section{Edge deployment}
\subsection{Deployment platform for edge devices}

\begin{figure}[hbt!]
\centering
\includegraphics[width=0.7\linewidth]{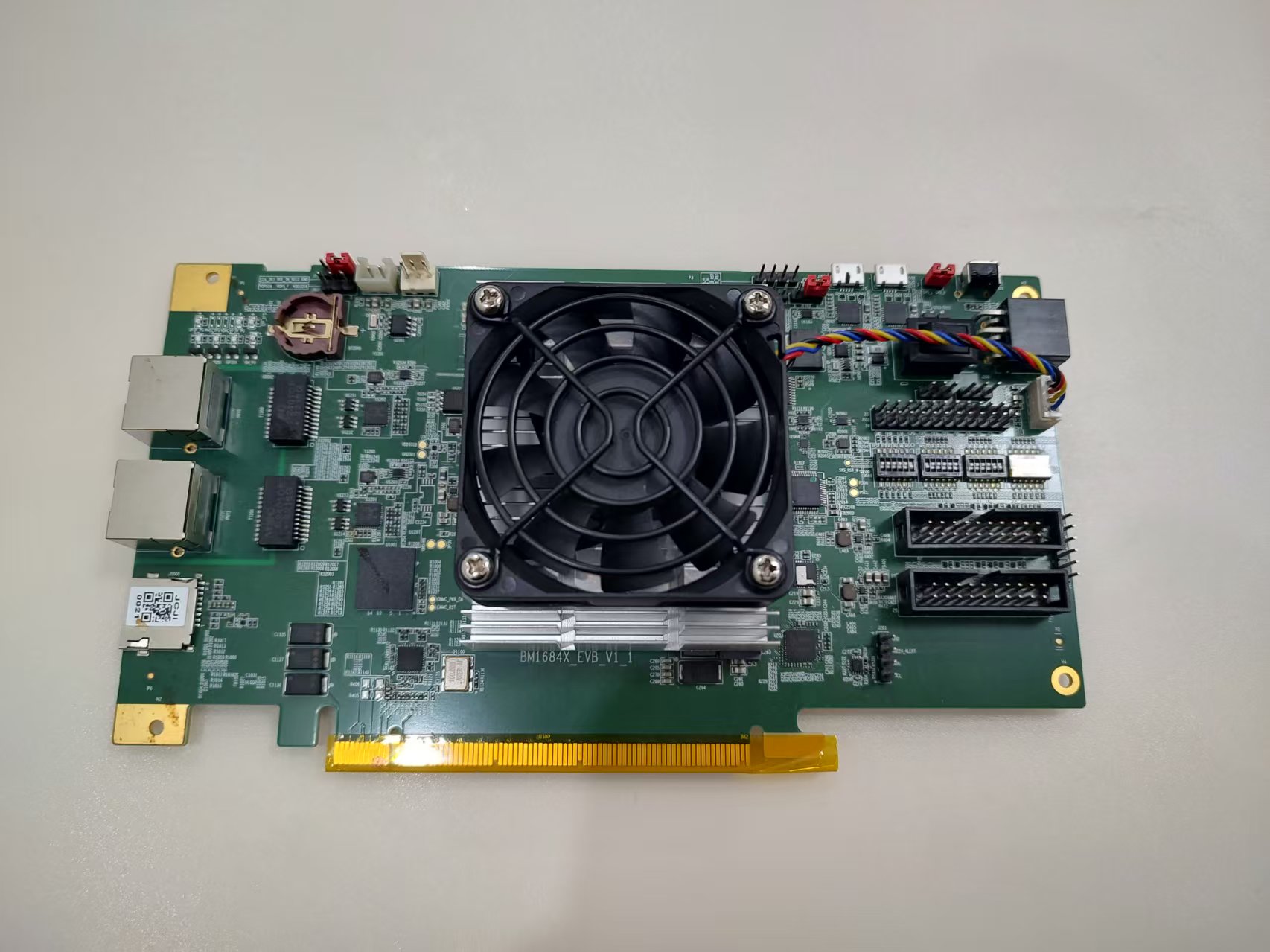}
\caption{Edge computing platform SOPHON BM1684X.}
\label{bm1684x}
\end{figure}

To verify the practical deployment performance of the proposed model on edge devices, this section deploys the model on the SOPHON BM1684X\autocite{bm1684x} edge computing platform. The device is shown in Figure \ref{bm1684x}, and its specifications are listed in Table \ref{cs}. This device utilizes a TPU (Tensor Processing Unit), a dedicated processor designed by Google specifically for machine learning tasks, aimed at accelerating the training and inference of deep learning models with high computational efficiency and low energy consumption.

\begin{table}[h]
    \centering
    \caption{Parameters of SOPHON BM1684X.}
    \begin{tabular}{p{0.35\textwidth}p{0.55\textwidth}}
        \toprule
        Parameter Name & Parameter Value \\
        \midrule
        TPU Architecture & Octa - core ARM Cortex - A53 @ 2.3GHz \\
        Peak Computing Power & INT8: 32 TOPS; FP16/BF16: 16 TFLOPS; FP32: 2 TFLOPS \\
        Supported Deep Learning Frameworks & TensorFlow, Caffe, PyTorch, Paddle, etc. \\
        Video Processing Capacity & Up to 32 channels of 1080P H.264/H.265 decoding \\
        Power Consumption & 20W \\
        \bottomrule
    \end{tabular}
\label{cs}
\end{table}

\subsection{Deployment Process and Experimental Verification}

\begin{figure}[hbt!]
\centering
\includegraphics[width=0.6\linewidth]{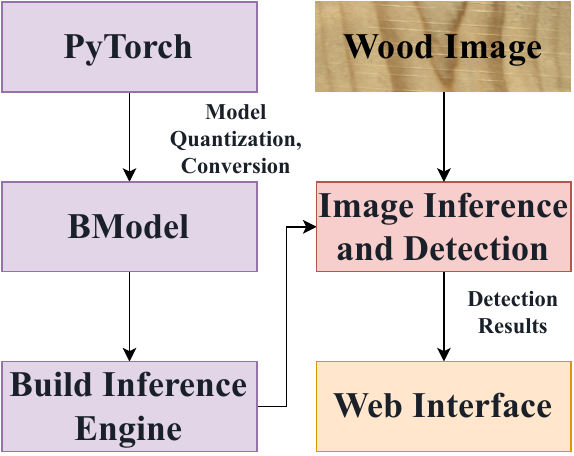}
\caption{The deployment process of CFIS-YOLO on the edge device.}
\label{deploy}
\end{figure}

\begin{figure*}[ht]
\centering
\includegraphics[width=\textwidth]{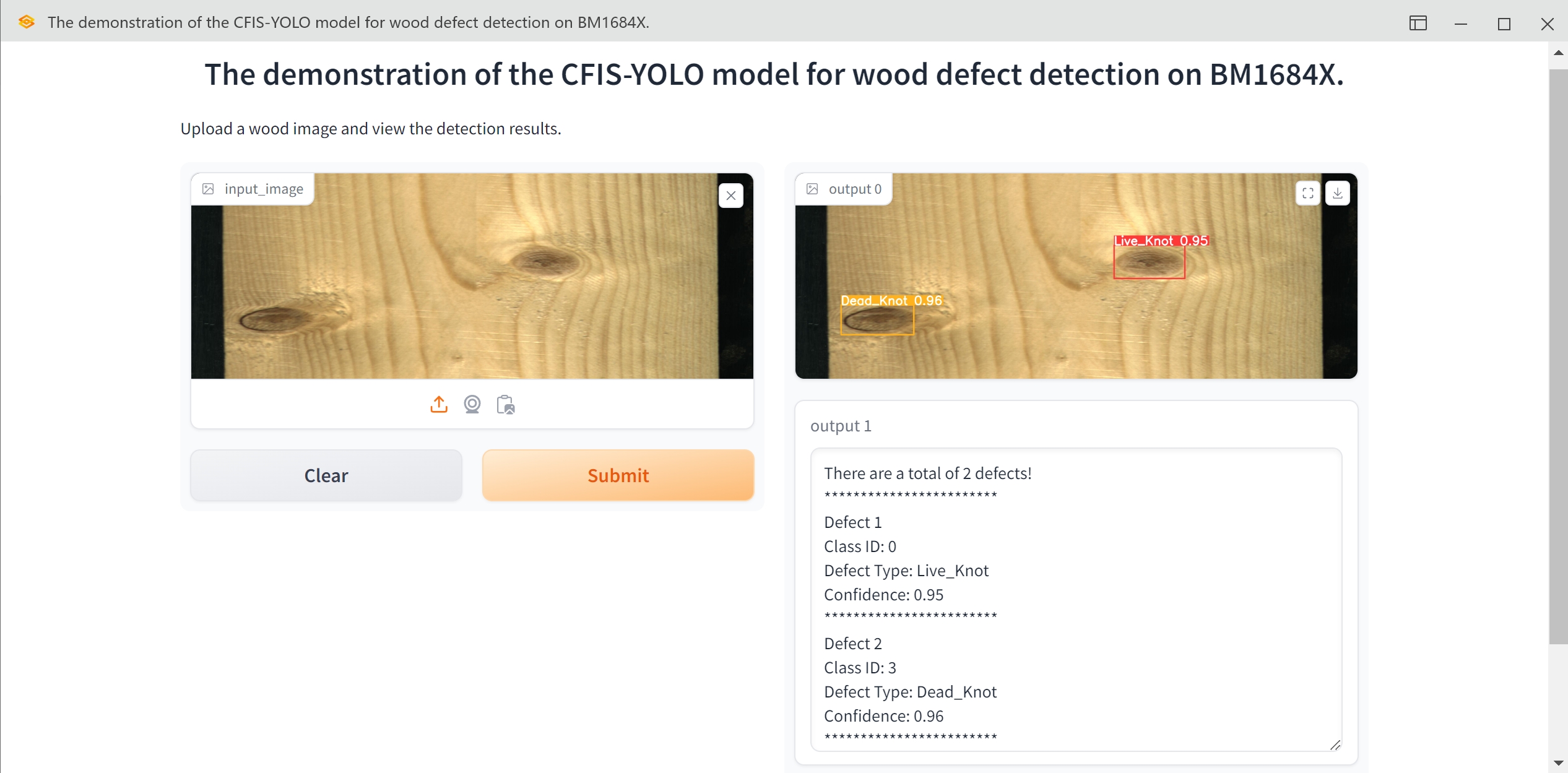}
\caption{Web demonstration interface of CFIS-YOLO.}
\label{web}
\end{figure*}

The deployment workflow of the wood defect detection algorithm is illustrated in Figure \ref{deploy}. For a PyTorch model trained on a GPU, the process involves model quantization, pruning, and conversion to generate a Bmodel format specifically adapted for BM1684X. Subsequently, an inference engine is created via the Bmodel to perform image inference and detection. To facilitate demonstration and user interaction, we employ Gradio, an open-source Python library for rapid web interface development, to construct a web page. Users can submit wood defect images to the edge device, and the detection results along with visualizations are displayed on the web interface. The demonstration interface for wood defect detection on BM1684X is shown in Figure \ref{web}.

Table \ref{bs} benchmarks CFIS-YOLO's performance across BM1684X and GPU. The results indicate that BM1684X achieves a slightly lower mAP50 than GPU due to fp16 quantization, which converts the original 32-bit floating-point model to 16-bit floating-point precision, resulting in minor accuracy degradation. However, BM1684X achieves a higher FPS of 135 frames per second, demonstrating significant advantages in inference speed. This is attributed to BM1684X's specialized optimization for deep learning inference tasks, particularly its support for low-precision computations and the lightweight improvements described in Chapter 2. Additionally, BM1684X exhibits significantly smaller model size and lower power consumption compared to GPU. In summary, compared to NVIDIA A10, BM1684X sacrifices only 0.5 percentage points in accuracy while achieving 17.3\% of GPU's power consumption, 78.3\% of the original PyTorch model size, and a 31\% FPS improvement, making it the optimal choice for deployment on edge computing devices or low-power scenarios.

\begin{table}[htbp]
\small
\centering
\setlength{\tabcolsep}{3pt}
\caption{Performance comparison of CFIS-YOLO on BM1684X and A10.}
\label{bs}
\begin{tabular}{ccccc}
\toprule
Device & mAP@0.5 (\%) & FPS & Model Size (MB) & Wattage (W) \\
\midrule
BM1684X & 77.0 & 135 & 11.21 & 17 \\
NVIDIA A10 & 77.5 & 103 & 14.3 & 98 \\
\bottomrule
\end{tabular}
\end{table}

\section{Conclusion}
This study addresses key challenges in wood defect detection, including insufficient multi-scale feature fusion, limited localization accuracy for small objects, and the trade-off between precision and efficiency in model deployment. We propose a lightweight and efficient object detection model, CFIS-YOLO. By innovatively introducing the CARAFE lightweight content-aware upsampling operator, FasterBlock module, and Inner-SIoU loss function, we significantly enhance model performance in wood defect detection tasks. Experimental results show that CFIS-YOLO achieves an mAP@0.5 of 77.5\% on public wood defect datasets, a 4 percentage point improvement over the baseline model YOLOv10s, demonstrating excellent detection accuracy. Deployment testing on the SOPHON BM1684X edge device further confirms the model's practicality, with power consumption reduced to 17.3\% of the original while sacrificing only 0.5 percentage points of mAP, fully meeting industrial demands for low power consumption and high real-time performance.
The innovations of this study are reflected in the following aspects: First, the CARAFE upsampling operator enhances semantic alignment of multi-scale features through dynamic kernel reorganization, significantly improving traditional upsampling methods for handling diverse wood textures. Second, the FasterBlock module reduces computational redundancy using partial convolutions, improving inference speed while maintaining feature representation, laying the foundation for edge deployment. Third, the Inner-SIoU loss function optimizes localization accuracy for small objects by integrating inner IoU and angle-aware mechanisms, effectively addressing the challenge of detecting tiny defects. These improvements collectively construct an efficient and lightweight detection framework, providing reliable technical support for the intelligent upgrading of the wood processing industry.
Despite significant achievements, this study has limitations. The current dataset's limited diversity and scale may restrict the model's generalization to broader scenarios. Additionally, the model's robustness under extreme conditions such as poor lighting or low image quality remains unverified. Future research directions include: optimizing the model structure to enhance detection accuracy and speed; expanding the dataset and testing deployment on more edge devices; and exploring applicability to other wood defect types to further improve practical value.
In summary, this study provides an efficient and lightweight solution for wood defect detection through multi-dimensional innovations in feature fusion, loss function design, and network optimization. The findings are not only academically significant, advancing object detection technology in specific industrial fields, but also offer a feasible technical path for the intelligent and sustainable development of the wood processing industry.

\paragraph{Disclosure statement}
No potential conflict of interest was reported by the author(s).

\printendnotes

\defbibnote{preamble}{By default, this template uses \texttt{biblatex} and adopts the Chicago referencing style. However, the journal you’re submitting to may require a different reference style; specify the journal you're using with the class' \texttt{journal} option --- see lines 1--8 of \emph{sample.tex} for a list of options and instructions for selecting the journal.}

\printbibliography

@ARTICLE{wlsd,
  author={Zhang, Qiyu and Liu, Liping and Yang, Ziyi and Yin, Jingtao and Jing, Zhizhong},
  journal={IEEE Access}, 
  title={WLSD-YOLO: A Model for Detecting Surface Defects in Wood Lumber}, 
  year={2024},
  volume={12},
  number={},
  pages={65088-65098},
  doi={10.1109/ACCESS.2024.3395623}}

@article{cwb4,
  title={Review of the Current State of Application of Wood Defect Recognition Technology.},
  author={Chen, Yutang and Sun, Chengshuo and Ren, Zirui and Na, Bin},
  journal={BioResources},
  volume={18},
  number={1},
  year={2023}
}

@article{fdd31,
  title={Economic efficiency of pine wood processing in furniture production},
  author={Wieruszewski, Marek and Turba{\'n}ski, Wojciech and Mydlarz, Katarzyna and Sydor, Maciej},
  journal={Forests},
  volume={14},
  number={4},
  pages={688},
  year={2023},
  publisher={MDPI}
}

@article{cwb,
  title={Wood defect detection based on the CWB-YOLOv8 algorithm},
  author={An, Hao and Liang, Zhihong and Qin, Mingming and Huang, Yuxiang and Xiong, Fei and Zeng, Guojian},
  journal={Journal of Wood Science},
  volume={70},
  number={1},
  pages={26},
  year={2024},
  publisher={Springer}
}

@article{bpn,
  title={BPN-YOLO: A Novel Method for Wood Defect Detection Based on YOLOv7},
  author={Wang, Rijun and Chen, Yesheng and Liang, Fulong and Wang, Bo and Mou, Xiangwei and Zhang, Guanghao},
  journal={Forests},
  volume={15},
  number={7},
  pages={1096},
  year={2024},
  publisher={MDPI}
}

@article{zou,
  title={An accurate object detection of wood defects using an improved Faster R-CNN model},
  author={Zou, Xianghe and Wu, Chongyang and Liu, Hongen and Yu, Zhangwei and Kuang, Xianyan},
  journal={Wood Material Science \& Engineering},
  pages={1--7},
  year={2024},
  publisher={Taylor \& Francis}
}

@INPROCEEDINGS{2019,
  author={Kurdthongmee, Wattanapong and Suwannarat, Korrakot},
  booktitle={2019 International Conference on Technologies and Applications of Artiﬁcial Intelligence (TAAI)}, 
  title={Locating Wood Pith in a Wood Stem Cross Sectional Image Using YOLO Object Detection}, 
  year={2019},
  volume={},
  number={},
  pages={1-6},
  doi={10.1109/TAAI48200.2019.8959823}}

@article{zheng2024,
  title={GBCD-YOLO: A high-precision and real-time lightweight model for wood defect detection},
  author={Zheng, Yunchang and Wang, Mengfan and Zhang, Bo and Shi, Xiangnan and Chang, Qing},
  journal={IEEE Access},
  volume={12},
  pages={12853--12868},
  year={2024},
  publisher={IEEE}
}

@ARTICLE{han2023,
  author={Han, Siyu and Jiang, Xiangtao and Wu, Zhenyu},
  journal={IEEE Access}, 
  title={An Improved YOLOv5 Algorithm for Wood Defect Detection Based on Attention}, 
  year={2023},
  volume={11},
  number={},
  pages={71800-71810},
  doi={10.1109/ACCESS.2023.3293864}}

@article{yolov5s,
title={Detection Model of Wood Surface Defects Based on Improved YOLOv5s},
author = {Zhu, Hao and Zhou, Shunyong and Zeng, Yalan and Li, Sicheng and Liu, Xue},
journal={Chinese Journal of Wood Science and Technology},
volume={37},
number={2},
pages={8-15},
year={2023},
doi={10.12326/j.2096-9694.2022163},
}

@article{wang2023,
  title={Tsw-yolo-v8n: Optimization of detection algorithms for surface defects on sawn timber},
  author={Wang, Mingtao and Li, Mingxi and Cui, Wenyan and Xiang, Xiaoyang and Duo, Huaqiong},
  journal={BioResources},
  volume={18},
  number={4},
  pages={8444},
  year={2023},
  publisher={North Carolina State University}
}

@article{zou2025,
  title={An improved method of AUD-YOLO for surface damage detection of wind turbine blades},
  author={Zou, Li and Chen, Anqi and Yang, Xinhua and Sun, Yibo},
  journal={Scientific Reports},
  volume={15},
  number={1},
  pages={5833},
  year={2025},
  publisher={Nature Publishing Group UK London}
}

@inproceedings{carafe,
  title={Carafe: Content-aware reassembly of features},
  author={Wang, Jiaqi and Chen, Kai and Xu, Rui and Liu, Ziwei and Loy, Chen Change and Lin, Dahua},
  booktitle={Proceedings of the IEEE/CVF international conference on computer vision},
  pages={3007--3016},
  year={2019}
}

@inproceedings{faster,
  title={Run, don't walk: chasing higher FLOPS for faster neural networks},
  author={Chen, Jierun and Kao, Shiu-hong and He, Hao and Zhuo, Weipeng and Wen, Song and Lee, Chul-Ho and Chan, S-H Gary},
  booktitle={Proceedings of the IEEE/CVF conference on computer vision and pattern recognition},
  pages={12021--12031},
  year={2023}
}

@article{PGY,
  title={Faster-PGYOLO: an efficient framework for floating debris detection in inland waters},
  author={Wang, Hongru and Cheng, Hu and Zhang, Jingtao},
  journal={The Visual Computer},
  pages={1--18},
  year={2024},
  publisher={Springer}
}

@article{siou,
  title={SIoU loss: More powerful learning for bounding box regression},
  author={Gevorgyan, Zhora},
  journal={arXiv preprint arXiv:2205.12740},
  year={2022}
}

@article{inner,
  title={Inner-iou: more effective intersection over union loss with auxiliary bounding box},
  author={Zhang, Hao and Xu, Cong and Zhang, Shuaijie},
  journal={arXiv preprint arXiv:2311.02877},
  year={2023}
}

@article{yolov10,
  title={Yolov10: Real-time end-to-end object detection},
  author={Wang, Ao and Chen, Hui and Liu, Lihao and Chen, Kai and Lin, Zijia and Han, Jungong and others},
  journal={Advances in Neural Information Processing Systems},
  volume={37},
  pages={107984--108011},
  year={2024}
}

@article{sjj,
  title={A large-scaleimage dataset of wood surface defects forautomated vision-based quality control processes},
  author={ Kodytek, Pavel.  and  Bodzas, Alexandra.  and  Bilik, Petr. },
  journal={F1000Research},
  volume={10},
  pages={581},
  year={2021},
}

@article{drr,
  title={DRR-YOLO: A Multiscale Wood Surface Defect Detection Method Based on Improved YOLOv8},
  author={Wang, Rijun and Chen, Yesheng and Zhang, Guanghao and Liang, Fulong and Mou, Xiangwei and Jin, Hao},
  journal={IEEE Sensors Journal},
  year={2025},
  publisher={IEEE}
}

@inproceedings{bm1684x,
  title={SOPHGO BM1684X: A Commercial High Performance Terminal AI Processor with Large Model Support},
  author={Gao, Peng and Liu, Yang and Wang, Jun and Cai, Wanlin and Shen, Guangchong and Hong, Zonghui and Qu, Jiali and Wang, Ning},
  booktitle={2024 57th IEEE/ACM International Symposium on Microarchitecture (MICRO)},
  pages={1413--1428},
  year={2024},
  organization={IEEE}
}

@article{ali2024yolo,
  title={The YOLO framework: A comprehensive review of evolution, applications, and benchmarks in object detection},
  author={Ali, Momina Liaqat and Zhang, Zhou},
  journal={Computers},
  volume={13},
  number={12},
  pages={336},
  year={2024},
  publisher={MDPI}
}

\end{document}